%% file: main.tex
\documentclass[runningheads]{llncs}

\usepackage{eccv}

\usepackage{eccvabbrv}

\usepackage{graphicx}
\usepackage{booktabs}

\usepackage[accsupp]{axessibility}  % Improves PDF readability for those with disabilities.

\usepackage{hyperref}

\usepackage{orcidlink}

\usepackage{pifont}
\usepackage{multirow}

\newcommand{\nbf}[1]{{\noindent \textbf{#1.}}}
\newcommand{\supp}{supplementary material\xspace}

\newcommand{\name}{D-SCo\xspace}

\newlength\savewidth
\newcommand{\tablestyle}[2]{\setlength{\tabcolsep}{#1}\renewcommand{\arraystretch}{#2}\centering\footnotesize}

\newcommand*\samethanks[1][\value{footnote}]{\footnotemark[#1]}
\newcommand{\cmark}{\ding{51}}
\newcommand{\xmark}{\ding{55}}

\begin{document}

\title{D-SCo: Dual-Stream Conditional Diffusion for Monocular Hand-Held Object Reconstruction} 

\titlerunning{D-SCo}


% ORCID
\author{Bowen Fu$^{1,2}$\orcidlink{0000-0002-0002-1165}, 
Gu Wang$^{1}$\orcidlink{0000-0002-0759-0782}, 
Chenyangguang Zhang$^{1}$\orcidlink{0009-0005-7852-4124}, 
Yan Di$^{2}$\orcidlink{0000-0003-0671-8323}, \\
Ziqin Huang$^{1}$\orcidlink{0009-0008-3104-2146}, 
Zhiying Leng$^{2}$\orcidlink{0000-0002-8773-6939}, 
Fabian Manhardt$^{3}$\orcidlink{0000-0002-4577-4590}, 
Xiangyang Ji$^{1}$\thanks{corresponding author}\orcidlink{0000-0002-7333-9975} \\
and Federico Tombari$^{2,3}$\samethanks\orcidlink{0000-0001-5598-5212} \\
\institute{$^1$Tsinghua University \quad $^2$Technical University of Munich \quad $^3$Google\\
\tt\small{\{fbw19, zcyg22, huang-zq20\}@mails.tsinghua.edu.cn}, \\
\{yan.di@, zhiying.leng@, tombari@in.\}tum.de, \\
fabianmanhardt@google.com, \{wanggu1, xyji\}@tsinghua.edu.cn
}
}

\authorrunning{B. Fu \etal}


\maketitle

\input{sec/0_abstract}    
\input{sec/1_intro}
\input{sec/2_related}
\input{sec/3_method}
\input{sec/4_exp}
\input{sec/5_conclusion}
\input{sec/6_acknowledgement}

\bibliographystyle{splncs04}
\bibliography{main}
\end{document}

%% file: sec/0_abstract.tex
\begin{abstract}

Reconstructing hand-held objects from a single RGB image is a challenging task in computer vision.
In contrast to prior works that utilize deterministic modeling paradigms, we employ a point cloud denoising diffusion model to account for the probabilistic nature of this problem. 
In the core, we introduce centroid-fixed \underline{D}ual-\underline{S}tream \underline{Co}nditional diffusion for monocular hand-held object reconstruction (\textbf{\name}), tackling two predominant challenges. 
First, to avoid the object centroid from deviating, we utilize a novel hand-constrained centroid fixing paradigm, enhancing the stability of diffusion and reverse processes and the precision of feature projection. 
Second, we introduce a dual-stream denoiser to semantically and geometrically model hand-object interactions with a novel unified hand-object semantic embedding, enhancing the reconstruction performance of the hand-occluded region of the object.
Experiments on the synthetic ObMan dataset and three real-world datasets HO3D, MOW and DexYCB demonstrate that our approach can surpass all other state-of-the-art methods. 

\keywords{Point Cloud Diffusion \and Hand-Held Object Reconstruction}

\end{abstract}

%% file: sec/1_intro.tex
\section{Introduction}
\label{sec:intro}

\input{figs/1_intro}

Modeling hand-object interaction has a wide range of applications across various domains, including AR/VR~\cite{leng2021stable, qian2022arnnotate, rantamaa2023comparison} and human-robot interaction~\cite{edsinger2007human, ortenzi2021object, gao2021dynamic, yang2021cpf}. 
Significant progress has been recently made for this task~\cite{leng2021stable, yang2021cpf, cao2021MOW}, where monocular hand-held object reconstruction draws particular attention~\cite{hasson_CVPR19_obman, karunratanakul_2020_gf, Ye_CVPR22_iHOI}. 
Reconstructing hand-held objects from a single RGB image is a highly challenging and ill-posed problem.
Suffering from the lack of real-world data and the ambiguity caused by hand- and self-occlusion, the performance of single-view hand-held object reconstruction remains limited \cite{hasson_CVPR19_obman,Ye_CVPR22_iHOI,zhang2024ddf}. 

In essence, most existing works either rely on Signed Distance Fields (SDFs)~\cite{karunratanakul_2020_gf, Ye_CVPR22_iHOI} or Directed Distance Fields (DDFs)~\cite{zhang2024ddf} to represent object shapes. 
Despite being effective under favorable conditions, such techniques generally tend to result in over-smoothed and undetailed reconstruction~\cite{park2019deepsdf,choe2021deep}.
Moreover, prior work~\cite{hasson_CVPR19_obman, karunratanakul_2020_gf, Ye_CVPR22_iHOI,zhang2024ddf} typically utilize a deterministic modeling paradigm, making it difficult to reason about the uncertainties introduced by hand- or self-occlusion.
Recently, probabilistic point cloud denoising diffusion models~\cite{luo2021diffusion, Zhou_21ICCV_PVD, Melas-Kyriazi_CVPR23_PC2} have shown to be effective for the task of single-view object reconstruction. 
Compared to works employing surface-based representations like SDFs and DDFs, diffusion-driven methods for reconstructing point clouds enjoy better capabilities in overcoming artifacts such as noise, sparsity, and irregularities~\cite{choe2021deep}. 
Moreover, the probabilistic nature of denoising diffusion models is particularly beneficial for modeling uncertainties and underconstrained problems.

Thus, in this work, we propose to leverage these probabilistic point cloud denoising diffusion models to conduct hand-held object reconstruction from a single RGB image. 
However, directly employing diffusion models in single-view hand-held object reconstruction faces two main problems: 
\textbf{First}, in existing diffusion models~\cite{Luo_21CVPR_DMPGen, xie2019pix2vox, xie2020pix2vox++}, the centroid of a partially denoised point cloud is not controlled and can thus deviate, for example, to the back side of the hand or even intersect with the hand, leading to physically implausible results. 
Additionally, the centroid deviation causes the misalignment of the semantic features~\cite{di2023ccd3dr} and can thus have adverse effects on the object reconstruction quality.  
\textbf{Second}, diffusion models for 3D reconstruction are typically conditioned only on single-stream 2D image features~\cite{xie2019pix2vox,Melas-Kyriazi_CVPR23_PC2}, without modeling geometric hand-object interaction or addressing the uncertainty induced by hand occlusion. 

To solve the aforementioned problems, we present \name, a centroid-fixed dual-stream conditional point cloud denoising diffusion model for single-view hand-held object reconstruction. 
As shown in \cref{fig:intro}, we compare our \name with naive diffusion models. 
First, a hand-constrained centroid fixing scheme is proposed to ensure that the centroid of the partially denoised point cloud does not diverge during the diffusion as well as the reverse process. 
In particular, we use a small and efficient neural network to estimate the hand-constrained object centroid and use it as a guide during the reverse process. 
Hence, the model only needs to consider the simpler shape diffusion task rather than having to simultaneously account for shape and position.
Second, to best leverage the hand-object interaction prior, we introduce a dual-stream denoiser, which individually processes semantic and geometric priors. 
In particular, we utilize a unified hand-object semantic embedding to compensate for hand occlusion in the 2D feature extraction stage. 
Our contributions can be summarized as follows. 

\begin{itemize}
\item We present \name, the first conditional point cloud diffusion model for 3D reconstruction of hand-held objects from a single RGB image.

\item We introduce a novel hand-constrained centroid fixing scheme, utilizing the hand vertices prior to prevent the centroid of the partially denoised point cloud from diverging during diffusion and reverse processes. 

\item We propose a dual-stream denoiser to semantically and geometrically model hand-object interactions. Our novel unified hand-object semantic embedding serves as a strong prior for reconstructing occluded regions of the object. 

\end{itemize}

\noindent 
Experiments on the synthetic ObMan~\cite{hasson_CVPR19_obman} dataset and three real-world datasets, \ie, HO3D~\cite{hampali_CVPR20_HO3D}, MOW~\cite{cao2021MOW} and DexYCB~\cite{dexycb}, demonstrate that \name can surpass all existing methods by a large margin.

%% file: figs/1_intro.tex
\begin{figure}
    \begin{center}
    \includegraphics[width=0.99\linewidth]{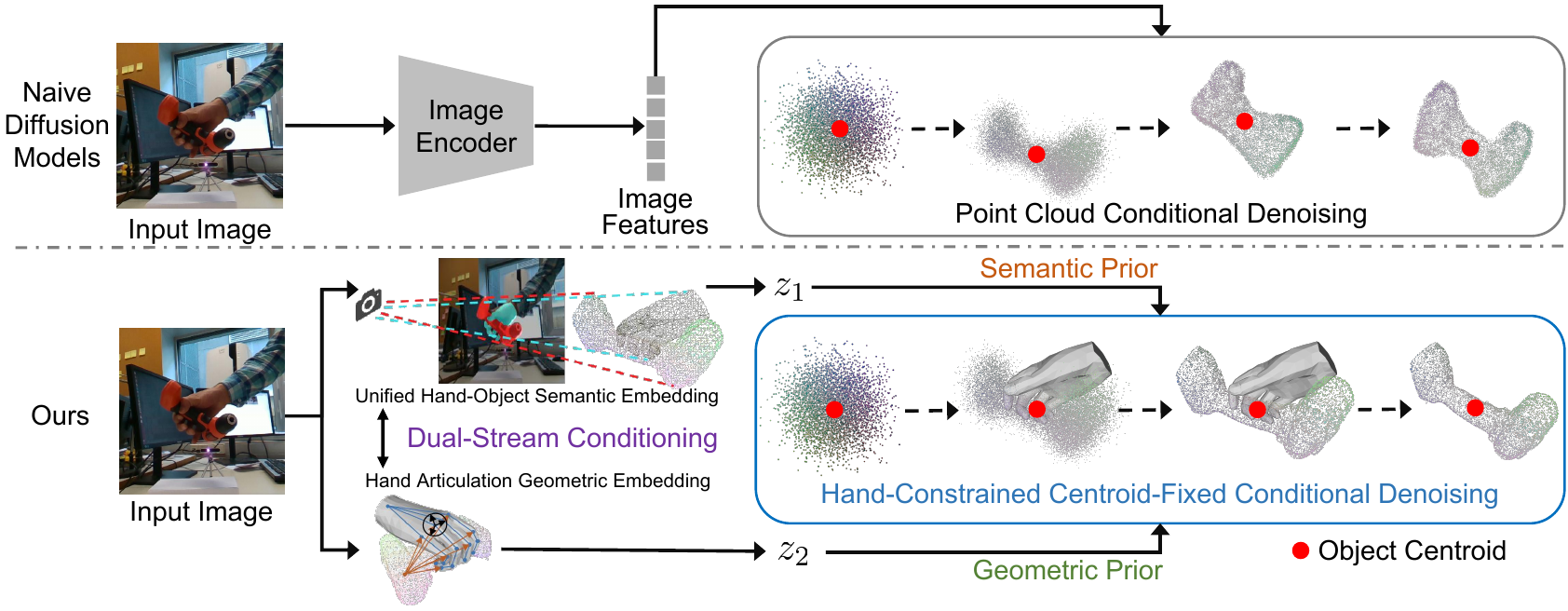}
    \end{center}
    \caption{
Comparison between \name and naive diffusion models for hand-held object reconstruction. \textbf{Naive diffusion models} are conditioned only on image features without controlling object centroid deviation or modeling the uncertainty induced by hand occlusion. 
\textbf{\name}, however, keeps the object centroid fixed under the constraint of the hand, making the diffusion model focus on shape reconstruction, and utilizes a dual-stream architecture to individually process semantic and geometric priors to learn a suitable representation for their own domain, tackling the aforementioned problems. 
}
\label{fig:intro}
\end{figure}

%% file: sec/2_related.tex
\section{Related Work}
\label{sec:related}

\nbf{Single-view Object Reconstruction}
Object reconstruction is a long-standing problem in the computer vision community. 
Traditionally, varieties of works focus on reconstruction from multi-view geometric techniques~\cite{hartley2003multiple,seitz2006comparison,schoenberger2016sfm,schoenberger2016mvs}. 
Recently, advanced by the powerful representation capacity of deep learning, single-view-based methods for object reconstruction have shown promising results~\cite{choy2016r2n2,xie2019pix2vox,xie2020pix2vox++}, despite the highly ill-posed setting.
Early learning-based methods propose to learn category-specific networks~\cite{kar2015category, kanazawa2018learning, tulsiani2016learning}, while more recent works try to learn a generalizable model across multiple categories for either meshes~\cite{groueix2018atlas, gkioxari2019mesh}, voxels~\cite{girdhar2016learning, choy2016r2n2, wu2016learning}, point clouds~\cite{fan2017point, lin2018learning}, or implicit representations such as NeRFs~\cite{mildenhall2020nerf} and SDFs~\cite{park2019deepsdf, chen2019learning}. 
In this work, we focus on the difficult yet very important problem of learning hand-held object reconstruction from a single RGB image~\cite{hasson_CVPR19_obman}, putting a particular emphasis on modeling the impact of the hand occlusion.

\nbf{Single-View Hand-Held Object Reconstruction}
Reconstructing hand-held objects is very challenging due to intricate hand- and self-occlusions.
Earlier works opt for the simplified task of 6DoF object pose estimation~\cite{tzionas2015_3d,sridhar2016real,Tekin_2019_CVPR_HO,hampali_CVPR20_HO3D,hasson2020leveraging}, assuming known object templates.
To better utilize hand-object interactions, some works jointly reason about hand and object poses by means of implicit feature fusion~\cite{chen2021joint,liu2021semi,shan2020understanding,Tekin_2019_CVPR_HO}, utilizing explicit geometric constraints~\cite{corona2020ganhand,brahmbhatt2020contactpose,grady2021contactopt,cao2021MOW}, or enforcing of physical constraints~\cite{tzionas2016capturing,pham2017hand}.
Increasing attention has recently been paid to model-free hand-held object reconstruction, as it is a more applicable setting.
Exemplarily, while \cite{hasson_CVPR19_obman} explicitly predicts the object mesh, \cite{karunratanakul_2020_gf,chen2022alignsdf,chen2023gsdf,zhang2024ddf} leverage implicit 3D representations such as SDFs, DDFs for reconstructing the shape of the hand and object. 
Recently, the concurrent work~\cite{zhang2024moho} instead utilizes NeRF as a representation. 
However, all these methods follow a deterministic approach, oftentimes resulting in low-quality reconstruction results for occluded and invisible parts, especially in the ill-posed single-view setting.
We instead draw inspiration from recent advances in probabilistic 3D generation~\cite{ho2020denoising,Zhou_21ICCV_PVD,di2023ccd3dr,Melas-Kyriazi_CVPR23_PC2} and additionally leverage semantic and geometric hand-object priors to achieve object reconstructions with high fidelity.

\nbf{Diffusion Models for 3D Reconstruction} 
Denoising diffusion models (DDMs) \cite{ho2020denoising,nichol2021improved} have recently attracted increasing attention in 3D reconstruction. 
While Luo \etal~\cite{Luo_21CVPR_DMPGen} and LION~\cite{zeng2022lion} employ latent variables for the diffusion process in point cloud generation,
PVD~\cite{Zhou_21ICCV_PVD} directly applies DDMs to point clouds, leading to a unified framework for both unconditional 3D generation and conditional completion of partial shapes.
Built upon PVD, PC$^2$~\cite{Melas-Kyriazi_CVPR23_PC2} conducts single-view reconstruction by conditioning on projected image features.
In this work, we employ point cloud diffusion models~\cite{Zhou_21ICCV_PVD,Melas-Kyriazi_CVPR23_PC2} for hand-held object reconstruction, as we observed that these models are more robust towards producing fragmented or distorted surfaces under ambiguous views than SDF-driven approaches~\cite{Ye_CVPR22_iHOI}.

\nbf{Hand Pose estimation}
Hand pose estimation methods from RGB(-D) images can be primarily categorized into model-free and model-based methods. 
Model-free approaches typically detect 2D keypoints, which are then lifted to 3D joint positions~\cite{rogez20153d, rogez2015understanding, zimmermann2017learning, mueller2018ganerated, panteleris2018using, iqbal2018hand, mueller2019real}, whereas model-based approaches commonly exploit statistical models such as MANO~\cite{2017MANO} to work in a low-dimensional parameter spaces~\cite{sridhar2015fast, boukhayma20193d, zhang2019end, zhou2020monocular, rong2021frankmocap}. 
Compared with model-free methods, the latter line of works exhibits better robustness to occlusions as well as domain discrepancies~\cite{Ye_CVPR22_iHOI}. 
Consequently, in this work, we rely on an off-the-shelf model-based approach~\cite{hasson_CVPR19_obman,rong2021frankmocap} to obtain the hand poses, which are then further leveraged within object shape inference. 

%% file: sec/3_method.tex
\section{Method}
\label{sec:method}

\input{figs/2_network}

Given a single RGB image $I$ capturing a hand-held object, we aim at reconstructing its shape as a 3D point cloud. 
As shown in \cref{fig:network}, we first utilize off-the-shelf methods~\cite{hasson_CVPR19_obman,rong2021frankmocap} to predict hand parameters $\phi_H$ and camera view $\phi_C$. 
Thereby, $\phi_H$ is defined with respect to the MANO~\cite{2017MANO} hand model, having 45DoF joint parameters, and $\phi_C$ represents the 6DoF pose of the hand wrist in the world reference system. 
Given the estimated hand vertices as obtained from the hand parameters, we employ a small yet efficient network to predict the object centroid $\widehat{\mathcal{M}}$. 
Eventually, we leverage the estimated centroid as part of the point cloud diffusion process to enable robust object point cloud reconstruction. 
Note that in contrast to existing diffusion methods, our model can thus fully focus on shape reconstruction, leading to improved performance. 
Furthermore, we introduce a dual-stream denoiser to first independently process and then aggregate semantic and geometric priors with a novel unified hand-object embedding, which helps the reconstruction of the hand-occluded regions of the object. 
Due to the probabilistic nature of diffusion models together with our explicit modeling of hand-object interaction, \name shows superior performance in handling uncertainties arising from hand- and self-occlusion. 

% In the following sections, we first introduce the fundamentals of conditional point cloud denoising diffusion models (\cref{sec:diffusion}). 
% We then explain our centroid-fixed conditional diffusion (\cref{sec:centroid}) before diving into our proposed dual-stream conditional denoiser (\cref{sec:conditioning}).  

\subsection{Conditional Point Cloud Denoising Diffusion}
\label{sec:diffusion}
We formulate single-view hand-held object reconstruction as conditional point cloud denoising diffusion, consisting of two Markov chains called the diffusion process and the reverse diffusion process.
Suppose that we have a target point cloud with $N$ points $X_0\in\mathbb{R}^{3N}$ from the conditional distribution $q(X|z)$, where $z=I_{\phi_C, \phi_H}$ is the input RGB image with the corresponding camera view $\phi_C$ and the hand pose $\phi_H$.
For the diffusion process, we gradually add Gaussian noise to the target point cloud at different levels $t\in\{1,\cdots, T\}$ as
\begin{equation}
q(X_t|X_{t-1}, z)=\mathcal{N}(X_t|z; \sqrt{1-\beta_t}X_{t-1}|z,\beta_t \mathbf{I}).
\end{equation}
Notice that with a fixed variance schedule $\{\beta_t\}_{t=0}^T$, $X_t$ can be simply expressed by $X_0$ according to
\begin{equation}
q(X_t|X_0, z) = \mathcal{N}(X_t|z; \sqrt{\bar{\alpha}_t}X_0|z, (1-\bar{\alpha}_t) \mathbf{I}),
\end{equation}
where $\alpha_t= 1 - \beta_t, \bar{\alpha}_t = \prod_{s=1}^t\alpha_s$. 
Therefore, we can reparameterize $X_t$ as a linear combination of $X_0$ and a noise variable $\epsilon \sim \mathcal{N}(\mathbf{0}, \mathbf{I})$ as follows
\begin{equation}
X_t = \sqrt{\bar{\alpha}_t}X_0 + \sqrt{1 - \bar{\alpha}_t}\epsilon.
\end{equation}

Starting with a point cloud sample $X_T$ from random Gaussian noise, the reverse process iteratively samples from $q(X_{t-1}|X_t,z)$ to remove the added noise from the diffusion process.
To approximate this reverse process, we train a point cloud conditional denoiser $\mathcal{D}_\theta(X_t, t, z)$ to learn the distribution $q(X_{t-1}|X_t,z)$ by 
\begin{equation} 
\begin{aligned}
p_\theta(X_{t-1}|X_t,z) = \mathcal{N}(X_{t-1}; \mu_\theta(X_t, t, z), \sigma_t^2\mathbf{I}),\\
\mu_\theta(X_t, t, z) = \frac{1}{\sqrt{\alpha_t}}(X_t - \frac{1 - \alpha_t}{\sqrt{1-\bar{\alpha}_t}}\epsilon_{\theta}(X_t, t, z)),
\end{aligned}
\end{equation}
where $\mu_\theta$ is the estimated mean.

\subsection{Hand-Constrained Centroid-Fixed Conditional Diffusion}
\label{sec:centroid}
Existing works~\cite{Ye_CVPR22_iHOI,zhang2024ddf} commonly obtain the camera view $\phi_C$ and the hand pose $\phi_H$ from standard hand pose estimation methods~\cite{hasson_CVPR19_obman, rong2021frankmocap}.
Without knowing the object pose during inference, we can only attempt to denoise the points in the hand wrist coordinate system. 
However, directly using a diffusion model to jointly learn the object's shape and pose can become very challenging~\cite{di2023ccd3dr}.
Thus, we propose a hand-constrained centroid fixing scheme to ease learning.

\nbf{Centroid-Fixed Diffusion}
We first reduce the problem of object pose estimation to centroid prediction, which defines a new object coordinate system. 
Essentially, while the origin of the object coordinate system is located at the centroid of the object, the axis orientation is shared with the hand wrist frame. 
Notice that during training, we can directly use the ground-truth object centroid $\mathcal{M}$ to constrain the object point cloud.
Therefore, we harness $\mathcal{M}$ to stabilize the diffusion process and make the point cloud fixed at $\mathcal{M}$. 
In particular, during the centroid-fixed diffusion, % (\cref{alg:sampling})
we re-center the object point cloud to $\mathcal{M}$ and guarantee that the added noise has zero-mean via 
\begin{equation}
\begin{aligned}
X_0 \sim q(X_0), &~X_0 \leftarrow X_0 - \bar{X_0} + \mathcal{M}, \\
\epsilon \sim \mathcal{N}(\mathbf{0}, &~\mathbf{I}), \epsilon \leftarrow \epsilon - \bar{\epsilon},
\end{aligned}
\end{equation}
where $\bar{X_0}$ denotes the centroid of $X_0$. 
Noteworthy, by keeping the object centroid unchanged, the misalignment error within the semantic feature projection due to centroid movements could also be alleviated. Thus, the training behavior becomes more stable as well.

\nbf{Centroid Prediction for Reverse Diffusion}
During the reverse diffusion process, we propose the use of a simple yet effective network $\mathcal{G}$ to estimate the translation of the object \wrt the hand wrist coordinate system.
Given the input RGB image $I$ along with the corresponding hand pose $\phi_H$ and camera view $\phi_C$, the predicted object centroid is obtained as
\begin{equation}
\widehat{\mathcal{M}} = \mathcal{G}(I_{\phi_C, \phi_H}). 
\end{equation}
As shown in \cref{fig:network} (II), $\mathcal{G}$ first encodes the hand vertices $X_H$ using a PointNet-like~\cite{qi2017pointnet} model to constrain the object centroid in 3D space. 
Subsequently, the global hand features are combined with the image features, extracted by a pre-trained ResNet-18~\cite{he2016resnet}, and processed by two parallel Multilayer Perceptrons (MLPs) to respectively output the 3D and 2D object centroid. 

During the reverse process% (\cref{alg:sampling})
, we instead start from $X_T \sim \mathcal{N}(\mathbf{0}, \mathbf{I}), X_T \leftarrow X_T - \bar{X}_T + \mathcal{\widehat{M}}$. 
We then re-center the predicted noise and restrict the centroid of the denoised point cloud at $\widehat{\mathcal{M}}$ to always remain locked at each step $t\in\{T-1, \cdots,0\} $ according to
\begin{equation}
\begin{aligned}
    \epsilon_\theta \leftarrow \epsilon_\theta - \bar{\epsilon}_\theta, \\
    X_{t} \sim p_{\theta}(X_{t}|X_{t+1}, z), &~X_{t} \leftarrow X_{t} - \bar{X}_{t} + \mathcal{\widehat{M}}.
\end{aligned}
\end{equation}
After the denoising process, we can directly obtain the reconstructed object point cloud by taking $X_0$.

Noteworthy, estimating object orientation is more challenging than predicting only the translation and will significantly increase the complexity of the network. 
Therefore, we only constrain the object centroid since previous works~\cite{Melas-Kyriazi_CVPR23_PC2, liu2019pvcnn} have shown diffusion models are able to reconstruct the object well even if it is not in a canonical orientation.

\subsection{Dual-Stream Conditional Point Cloud Denoising}
\label{sec:conditioning}
The essence of the conditional point cloud denoiser $\mathcal{D}_\theta(X_t, t, z)$ is the modeling of the condition $z$ so to fully utilize the information of the input image along with the corresponding camera view and hand pose. 
Our key insight is that the hand can offer both semantic and geometric priors to facilitate object reconstruction. 
Therefore, we utilize a dual-stream architecture to individually process these priors to avoid mutual interference. 

\nbf{Unified Hand-Object Semantic Embedding}
Related work has shown that using 2D image features is a vital component for object reconstruction~\cite{xie2019pix2vox,xie2020pix2vox++}. 
Unlike prior works that extract a global embedding for the image~\cite{choy2016r2n2, xie2019pix2vox, xie2020pix2vox++}, we instead extract unique deep image features for each point of the partially denoised point cloud at each diffusion step in a projective manner. 
Specifically, we first extract image features $\mathcal{F}\in\mathbb{R}^{H\times W\times C}$ using a standard 2D network, such as ResNet~\cite{he2016resnet} or ViT~\cite{dosovitskiy2020vit}, with $C$ being the number of output feature channels. 
We then project the point cloud onto the image by the efficient point cloud rasterizer~\cite{ravi2020pytorch3d} $\mathcal{R} (\phi_C, X_t)$. 
Subsequently, the pointwise semantic features at timestep $t$ can be represented as 
$X_t^O = \pi(\mathcal{R} (\phi_C, X_t), \mathcal{F})$ with $X_t^O\in \mathbb{R}^{N\times C}$, where $\pi(\cdot)$ denotes the back-projection to 3D space. 
In this way, every individual point is pixel-aligned with the deep features corresponding to the pixel onto which the point is rasterized. 

However, due to the inevitable occlusion induced by the hand, it is very challenging to reconstruct the occluded part of the object as only a single view is provided.
On the other side, the hand information should not be fully ignored as the hand naturally serves as a prior for estimating the shape of the hand-occluded object region. 
Therefore, we apply $\pi( \mathcal{R(\cdot)})$ to both object and hand points to obtain $X_t^{HO} = \pi( \mathcal{R} (\phi_C, [X_t, X_{H}]), \mathcal{F})$ with $X_t^{HO}  \in \mathbb{R}^{(N + N_H)\times C}$, serving as a unified hand-object semantic embedding. 
Thereby, $N_H$ denotes the number of the hand mesh vertices $X_H$, which can be obtained from the estimated hand pose $\phi_H$ together with the MANO hand model~\cite{2017MANO}. 
Noteworthy, we also apply an extra one-hot encoding to indicate whether the points belong to the object or the hand, making $X_t^{HO} \in \mathbb{R}^{(N + N_H)\times (C+1)}$. 

Compared with existing works~\cite{Ye_CVPR22_iHOI,zhang2024ddf}, our proposed unified hand-object semantic embedding $X_t^{HO}$ holds information of the hand-induced occlusion, which is crucial for the reconstruction of the occluded part of the object, thus increasing the robustness (see \cref{tab:sota_obman} (b)). 

\nbf{Hand Articulation Geometric Embedding}
Furthermore, the object shape is also highly constrained by the hand and very geometrically related to the hand articulation. 
Inspired by~\cite{Ye_CVPR22_iHOI, zhang2024ddf, chen2023gsdf}, we explicitly encode hand-object interaction by transforming the partially denoised point cloud at every step to each hand joint frame.

Given the partially denoised point cloud $X_t$ and hand parameters $\phi_H$, we first compute the rotation $R_j$ and translation $T_j$ of each joint $j$ with respective to the hand wrist using forward kinematics given the hand model. 
We then transform $X_t$ from the hand wrist coordinate system to each hand joint frame via $X_t^j = R_jX_t + T_j \in \mathbb{R}^{N\times 3}$ to encode the hand articulation onto object points. 
Finally, the hand articulation embedding $X_t^A \in \mathbb{R}^{N\times J}$ is calculated via concatenation and flattening of $X_t^j$, with $J$ equaling 45 in our experiments as 15 hand joints are utilized. 

\nbf{Dual-Stream Denoiser}
The naive way to utilize the aforementioned semantic and geometric priors is to directly concatenate them as the condition $z = [X_t^{HO}, X_t^A]$ and produce the denoised point cloud controlled by $z$ in a single-stream manner. 
However, forcibly integrating these embeddings from different domains may result in interference and cause a drop in performance (see \cref{tab:ablation}).

Hence, we instead employ a dual-stream denoiser to separately process $X_t^{HO}$ and $X_t^A$. 
As shown in \cref{fig:network}, given $z_1 = X_t^{HO}$ and $z_2 = X_t^A$, we first feed the object and hand points along with their corresponding semantic embedding 
% $[X_t, z_1] \in \mathbb{R}^{(N + N_H)\times (C+4)}$ to $f_{\theta}^1$
$[X_t, z_1]$ to one branch of the dual-stream denoiser $f_\theta^1$ and obtain $\mathcal{F}_\theta^1 \in \mathbb{R}^{(N + N_H)\times S}$ as feature representation guided by the semantic prior, with $S$ denoting the number of latent feature channels.
Similarly, we feed the object point cloud along with its corresponding geometric embedding $[X_t, z_2]$ to $f_\theta^2$ using the identical architecture to obtain $\mathcal{F}_\theta^2 \in \mathbb{R}^{N\times S}$ as our geometric feature. 
The final noise is then predicted from the concatenation of the semantic and geometric features with $\epsilon_\theta = g_\theta([\mathcal{F}_\theta^1, \mathcal{F}_\theta^2]) \in \mathbb{R}^{N\times 3}$, where $g_\theta$ consists of stacked MLPs. 

In this way, each branch can learn a specialized representation suitable for its own domain and then serve as conditioning to the diffusion model, contributing to the reconstruction of object shape from different domains. 
The detailed architecture of the dual-stream denoiser is provided in \supp. 

\input{figs/vis}
\subsection{Training Objectives}

\nbf{Diffusion model}
For optimization, we use the common MSE loss between model prediction $\epsilon_\theta(X_t, t, z)$ and applied noise $\epsilon$ with
\begin{equation}
\label{eq:loss_denoise}
\mathcal{L}_{denoise} = \|\epsilon - \epsilon_\theta(X_t, t, z) \|, \epsilon \sim \mathcal{N}(\mathbf{0}, \mathbf{I}) . 
\end{equation}
We further regularize the object shape using a projective mask loss $\mathcal{L}_{mask}$. 
To this end, at each timestep $t$ the sampled and predicted point cloud are projected onto the image via the aforementioned rasterizer, and the L1 loss between them is computed as
\begin{equation}
\mathcal{L}_{mask} = \|\mathcal{R}(X_t) - \mathcal{R}(\hat{X_t})\|_1,
\end{equation}
where $\hat{\bullet}$ denotes predicted results. 
The overall loss is then a weighted sum of both terms with
\begin{equation}
\mathcal{L}_{overall} = \mathcal{L}_{denoise} + \eta_1 \mathcal{L}_{mask},  
\end{equation}
where the weight $\eta_1$ controls the strength of the projective 2D regularization. 

\nbf{Centroid prediction network}
Notice that the centroid prediction network is separately trained from $\mathcal{D}_\theta$. Thereby, the 3D object centroid $\mathcal{M}_{3d}$ in the hand wrist frame and the 2D object centroid $\mathcal{M}_{2d}$ in the normalized device coordinate (NDC) space are both supervised along with a 2D-3D projection loss. 
Overall, our total loss is defined as
\begin{equation}
\begin{split}
\mathcal{L}_{centroid} = \|\mathcal{M}_{3d} - \widehat{\mathcal{M}}_{3d} \| + \lambda_1 \|\mathcal{M}_{2d} - \widehat{\mathcal{M}}_{2d} \| + \lambda_2 \|\mathcal{P}(\widehat{\mathcal{M}}_{3d}) - \widehat{\mathcal{M}}_{2d} \|,
\end{split}
\end{equation}
where $\lambda_1$ and $\lambda_2$ are hyperparameters, $\mathcal{P}$ is the transformation from the hand wrist frame to the NDC space, and $\widehat{\bullet}$ refers to predicted results.

%% file: figs/2_network.tex
\begin{figure*}[t]
    \begin{center}
    \includegraphics[width=0.99\linewidth]{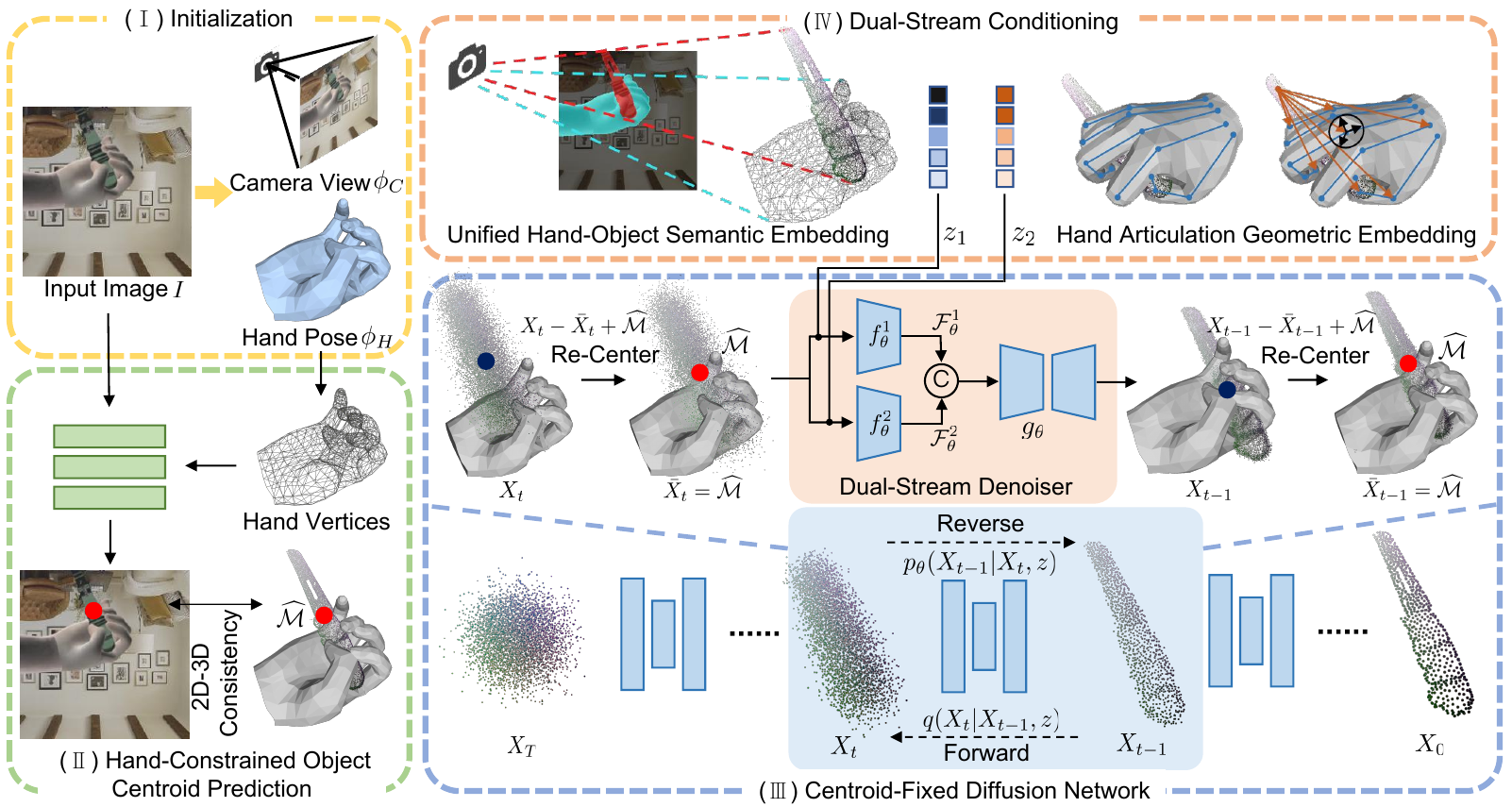}
    \end{center}
    \caption{
        \label{fig:network}
        \textbf{Architecture of \name.}
        (\uppercase\expandafter{\romannumeral1}) Given a single-view RGB image, we first predict the hand pose $\phi_H$ and camera view $\phi_C$ by an off-the-shelf network. 
        (\uppercase\expandafter{\romannumeral2}) The object centroid $\widehat{\mathcal{M}}$ is then estimated by our simple yet efficient hand-constrained centroid prediction network. 
        (\uppercase\expandafter{\romannumeral3}) We further introduce a centroid-fixed diffusion network, which always keeps the centroid of partially denoised point cloud fixed at the predicted centroid $\widehat{\mathcal{M}}$ during the reverse process. 
        (\uppercase\expandafter{\romannumeral4}) A dual-stream denoiser is proposed to individually process and then aggregate semantic and geometric hand-object interaction priors as condition. A unified hand-object semantic embedding is introduced to serve as a strong prior of hand-occlusion. 
        }

\end{figure*}

%% file: figs/vis.tex
\begin{figure*}[t]
    \begin{center}
    \includegraphics[width=0.99\linewidth]{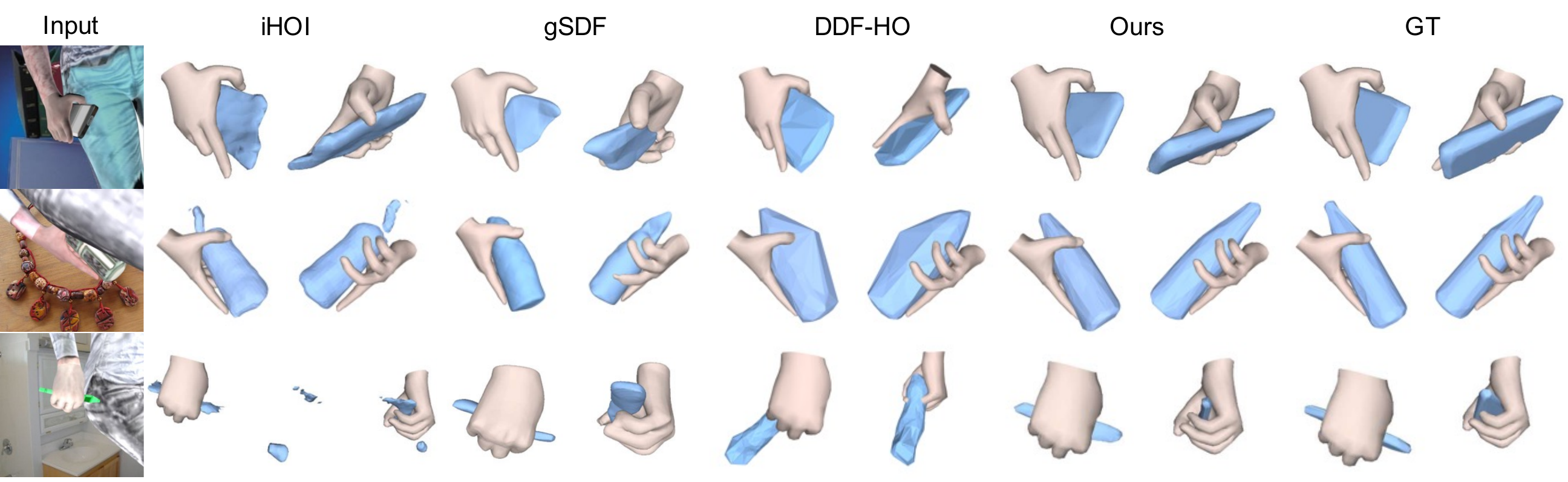}
    \end{center}
    \caption{
        \label{fig:vis}
        {Qualitative results on the ObMan~\cite{hasson_CVPR19_obman} dataset. For each method and ground truth, we show the reconstruction results in the camera view (column 1) and a novel view (column 2). }
    }

\end{figure*}

%% file: sec/4_exp.tex
\section{Experiments}
\label{sec:exp}

\subsection{Datasets and Setup} 
\nbf{Datasets}
We evaluate our method on four common benchmark datasets, including the synthetic ObMan~\cite{hasson_CVPR19_obman} dataset and three real-world datasets, namely HO3D~\cite{hampali_CVPR20_HO3D}, MOW~\cite{cao2021MOW} and DexYCB~\cite{dexycb}. 
\textbf{ObMan}~\cite{hasson_CVPR19_obman} consists of 2,772 objects from 8 categories taken from ShapeNet~\cite{chang2015shapenet}, with 21K plausible grasps generated using GraspIt~\cite{miller2004graspit}. 
The dataset contains 141K frames for training and 6K frames for testing, which are rendered using Blender on top of random backgrounds. 
\textbf{HO3D}~\cite{hampali_CVPR20_HO3D} includes 77,558 frames from 68 sequences, containing 10 different objects from the YCB dataset~\cite{calli2015ycb}, which are manipulated by 10 different users.
The hand and object pose annotations are obtained from multi-camera optimization procedures. 
\textbf{MOW}~\cite{cao2021MOW} consists of 442 images of 121 object templates, with in-the-wild source data collected from EPIC Kitchens~\cite{damen2018scaling, Damen2021PAMI} and the 100 Days of Hands~\cite{shan2020understanding} datasets. 
We use the same training and testing split as iHOI~\cite{Ye_CVPR22_iHOI} and DDF-HO~\cite{zhang2024ddf} for all three aforementioned datasets. 
\textbf{DexYCB}~\cite{dexycb} is a large-scale real-world hand-object dataset. 
Following \cite{chen2022alignsdf,yang2022artiboost,chen2023gsdf}, we focus on right-hand samples and use the official s0 split. 
29,656 training samples and 5,928 testing samples are downsampled following the setup of gSDF\cite{chen2023gsdf}. 

\input{tables/SOTA_obman}
\nbf{Metrics}
Following \cite{Ye_CVPR22_iHOI, zhang2024ddf}, we report the Chamfer Distance (CD) in mm and F-score at thresholds of 5mm (F-5) and 10mm (F-10). % to compare with the state-of-the-art.

% \noindent \textbf{Implementation Details} are provided in \supp. 

\subsection{Evaluation on ObMan Dataset}

We first present quantitative results on the large-scale synthetic ObMan dataset in \cref{tab:sota_obman} (a).  
As shown, our approach surpasses all state-of-the-art methods by a large margin for F-score and Chamfer Distance. 
Specifically, compared with the current best-performing method DDF-HO~\cite{zhang2024ddf}, we achieve a relative improvement of 10.9~\% and 20.9~\% for F-5 and F-10 metrics, respectively. 
Furthermore, we reduce the CD by 21.4~\% compared with DDF-HO and 89.2~\% compared with iHOI, demonstrating that our reconstructed shape possesses significantly fewer outliers. 

We also visualize the reconstructed objects along with the hands in \cref{fig:vis}. 
Since we focus on object reconstruction, we use ground-truth hand poses for all images. 
The object surfaces are reconstructed using alpha shapes~\cite{edelsbrunner1983alphashape} implemented with Open3D~\cite{zhou2018open3d} to qualitatively compare with previous works.
Noteworthy, the metrics are not affected as they are computed from the point clouds. 
While SDF- or DDF-based methods, including iHOI~\cite{Ye_CVPR22_iHOI}, gSDF~\cite{chen2023gsdf} and DDF-HO~\cite{zhang2024ddf}, tend to result in either over-smoothed and less-detailed, or fragmented reconstructions, our approach is able to generate geometrically coherent point clouds with plausible details, even for thin objects and heavily occluded parts. 
Specifically, the bottle in Row 2 suffers a 54.4\% occlusion rate. 
Nonetheless, \name shows strong capabilities of inferring the occluded part of the object. 

\input{tables/real_finetune_zeroshot}
\footnotetext{The results are provided by the authors of DDF-HO~\cite{zhang2024ddf}, with exact same training split as iHOI~\cite{Ye_CVPR22_iHOI} and ours. \label{footnote:ddfho}}

\subsection{Evaluation on Real-World Datasets} 
Aside from the synthetic ObMan dataset, we also conduct experiments on the three real-world datasets HO3D~\cite{hampali_CVPR20_HO3D}, MOW~\cite{cao2021MOW} and DexYCB~\cite{dexycb} to demonstrate our approach's generalization capabilities for real-world scenarios. 
The model is respectively finetuned on the HO3D and MOW, starting from the ObMan pre-trained model as initialization. 
As shown in \cref{tab:real_fintune_zeroshot} (top) and \cref{tab:sota_dexycb}, our approach achieves state-of-the-art results on HO3D, MOW as well as DexYCB. 
Concretely, regarding F-5 and F-10 metrics, we exhibit a noticeable average improvement of 54.7~\%, 86.2~\%, and 29.4~\%, respectively. 
In contrast to the F-score metric, the Chamfer Distance is known to be more vulnerable to outliers~\cite{tatarchenko2019single, Ye_CVPR22_iHOI, chen2023gsdf}. 
Thus, the significant reduction of the CD metric illustrates that our approach is significantly more robust, producing fewer outliers. 
The qualitative results in \cref{fig:vis_real} similarly demonstrate the superiority of our approach. 

\nbf{Zero-shot transfer to HO3D and MOW}
To further evaluate our zero-shot transfer abilities, we also directly apply our model, trained on ObMan, to HO3D and MOW without conducting any additional finetuning. 
The results in \cref{tab:real_fintune_zeroshot} (bottom) show that our method again achieves a remarkable improvement on HO3D (13~\% and 14~\% in F-5 and F-10) and MOW (71~\% and 89~\% in F-5 and F-10), demonstrating \name's strength in synthetic-to-real generalization. 

\input{tables/dexycb}
\subsection{Ablation Studies}
\nbf{Effectiveness of dual-stream denoiser}
We demonstrate the effectiveness of dual-stream denoiser in \cref{tab:ablation}. 
Instead of individually processing semantic and geometric embeddings, we simply concatenate the embeddings as conditioning to the denoiser. 
The decrease in performance (C0 \vs B0) shows that independently processing semantic and geometric information within our dual-stream denoiser enhances reconstruction quality. 

\nbf{Effectiveness of semantic and geometric condition}
In \cref{tab:ablation}, we show the impact of the semantic and geometric embeddings. 
The proposed unified hand-object semantic embedding $X_t^{HO}$ utilizes the semantic information of the hand to supplement the occluded semantic information of objects, which is crucial to model hand-induced occlusion. 
Adding the embedding leads again to improved performance (D2 \vs D0, C0 \vs D1). 
Moreover, the hand articulation geometric embedding $X_t^A$ explicitly models hand-object interaction geometrically and results in improved performance (D1 \vs D0, C0 \vs D2). 

We further noticed that the way of encoding the hand-object conditional information has a significant impact on performance.
In particular, we have explored two alternative strategies to model the hand-object interaction. 
% 1
First, inspired by~\cite{doosti2020hopenet}, we implement a GCN-based hand embedding, which utilizes a graph convolutional network~\cite{gao2019graph} for feature extraction. 
We again apply the GCN hand embedding to the partially denoised point cloud at each step.
Although the GCN hand embedding implicitly encodes the hand articulation, it does not encode the hand-object interaction. 
Therefore, simply applying such an embedding as conditioning does not further performance (See D3 \vs C0). 
The hand articulation geometric embedding, on the other hand, applies the unique articulation-aware embedding to each point of the partially denoised point cloud, which explicitly encodes hand-object interaction, 
thus significantly benefiting the object reconstruction performance. 
% 2
Second, we also utilize standard PointNet~\cite{qi2017pointnet} to encode 3D hand vertices into a global hand embedding, which is then served to the partially denoised point cloud. 
Again, without modeling the hand-object interaction, the global hand embedding falls short of providing sufficient semantic information about the hand-occluded object. 
Consequently, the predictions end up being inferior to our proposed $X_t^{HO}$ (D4 \vs C0). 

\input{figs/vis_real}
\nbf{Effectiveness of hand-constrained centroid fixing}
In \cref{tab:ablation}, we illustrate the importance of the centroid fixing scheme. 
Our centroid fixing operation improves the stability of the diffusion and the reverse processes, thus enhancing the performance (D0 \vs E0).
Further, without the hand-constrained centroid prediction network, the diffusion model has to simultaneously learn the centroid deviation and object shape, which leads to clearly worse results (E1 \vs D0). 
Additionally, in F0 we report results when using the actual ground-truth object centroid, and in F1 when using the ground-truth object pose. 
Without our centroid fixing scheme, the results remain inferior even when using the ground-truth object centroid on ObMan (F0 \vs D0). 
This further demonstrates the power of our proposed centroid fixing paradigm. 

\nbf{Effectiveness of $\mathcal{L}_{mask}$}
We compare the quantitative results w/ or w/o $\mathcal{L}_{mask}$. 
Supervising the diffusion process in both 2D and 3D domains further boosts the object reconstruction performance (A0 \vs B0). 
Noteworthy, we can surpass existing methods on both ObMan and HO3D datasets even without $\mathcal{L}_{mask}$ (B0). 

\input{tables/ablation}

\nbf{Robustness against occlusion}
To illustrate the robustness of our approach against hand occlusion, we split the test set of ObMan into groups according to the visible ratio of the object and compute the mean F-score for each group. 
As shown in \cref{tab:sota_obman} (b), when the object is undergoing strong occlusion by the hand (object visible ratio $<$ 50\%), iHOI suffers a significant decline in both F-5 and F-10 metrics. 
Similarly, DDF-HO also experiences an apparent decrease with $<$ 60~\% visible ratio. 
Nevertheless, the performance of our approach instead remains high. 
Aided by our hand-constrained centroid fixing and the modeling of the hand-induced occlusion, \name exhibits strong robustness against occlusion. 

\nbf{Oracle Experiments}
In line with other diffusion models~\cite{luo2021diffusion, Melas-Kyriazi_CVPR23_PC2} for 3D object reconstruction, in~\cref{tab:ablation} (A1) we report oracle results for \name. 
To this end, we predict five possible shapes for each object starting from different sampled Gaussian noises and choose the best sample for each input image with respect to the F-score. Essentially, the oracle results are supposed to demonstrate the probabilistic nature of our method and represent an upper bound.  
The obtained results underline the ability of \name to overcome the ill-posed essence of the problem. 
Thanks to the probabilistic formulation in diffusion models, our approach is able to generate multiple plausible shapes (See \supp), illustrating our capability of modeling the uncertainty induced by hand- and/or self-occlusion. 

% The robustness against hand pose prediction quality is further discussed in \supp. 

%% file: tables/SOTA_obman.tex
\begin{table}[t]
\centering
\tablestyle{15pt}{1.0}
\caption{
\label{tab:sota_obman}
{Comparison with the state-of-the-art on ObMan~\cite{hasson_CVPR19_obman}. 
(a) F-score of 5mm and 10mm, Chamfer Distance (mm) metrics are utilized for evaluation. 
(b) Robustness against hand occlusion. We analyze the patterns of the F-5 and F-10 metrics as a function of the object visibility ratio on the test set of ObMan. }
}
\begin{subfigure}[t]{0.48\linewidth}
\vspace{0pt}
\tablestyle{8pt}{0.955}
\scalebox{0.88}{
\begin{tabular}{@{}cccc@{}}
\toprule
Method & F-5 $\uparrow$ & F-10 $\uparrow$ & CD $\downarrow$ \\
\cmidrule(lr){2-4}
HO~\cite{hasson_CVPR19_obman} & 0.23 & 0.56 & 0.64 \\
GF~\cite{karunratanakul_2020_gf} & 0.30 & 0.51 & 1.39 \\
AlignSDF~\cite{chen2022alignsdf} & 0.40 & 0.64 & - \\
iHOI~\cite{Ye_CVPR22_iHOI} & 0.42 & 0.63 & 1.02 \\
gSDF~\cite{chen2023gsdf} & 0.44 & 0.66 & - \\
DDF-HO~\cite{zhang2024ddf} & 0.55 & 0.67 & 0.14 \\
Ours & \textbf{0.61} & \textbf{0.81} & \textbf{0.11} \\
\bottomrule
\end{tabular}
}
\caption{\label{fig:obman_results}}
\end{subfigure}
%%%%%%%%%%
\begin{subfigure}[t]{0.49\linewidth}
\vspace{0pt}
\includegraphics[width=\linewidth]{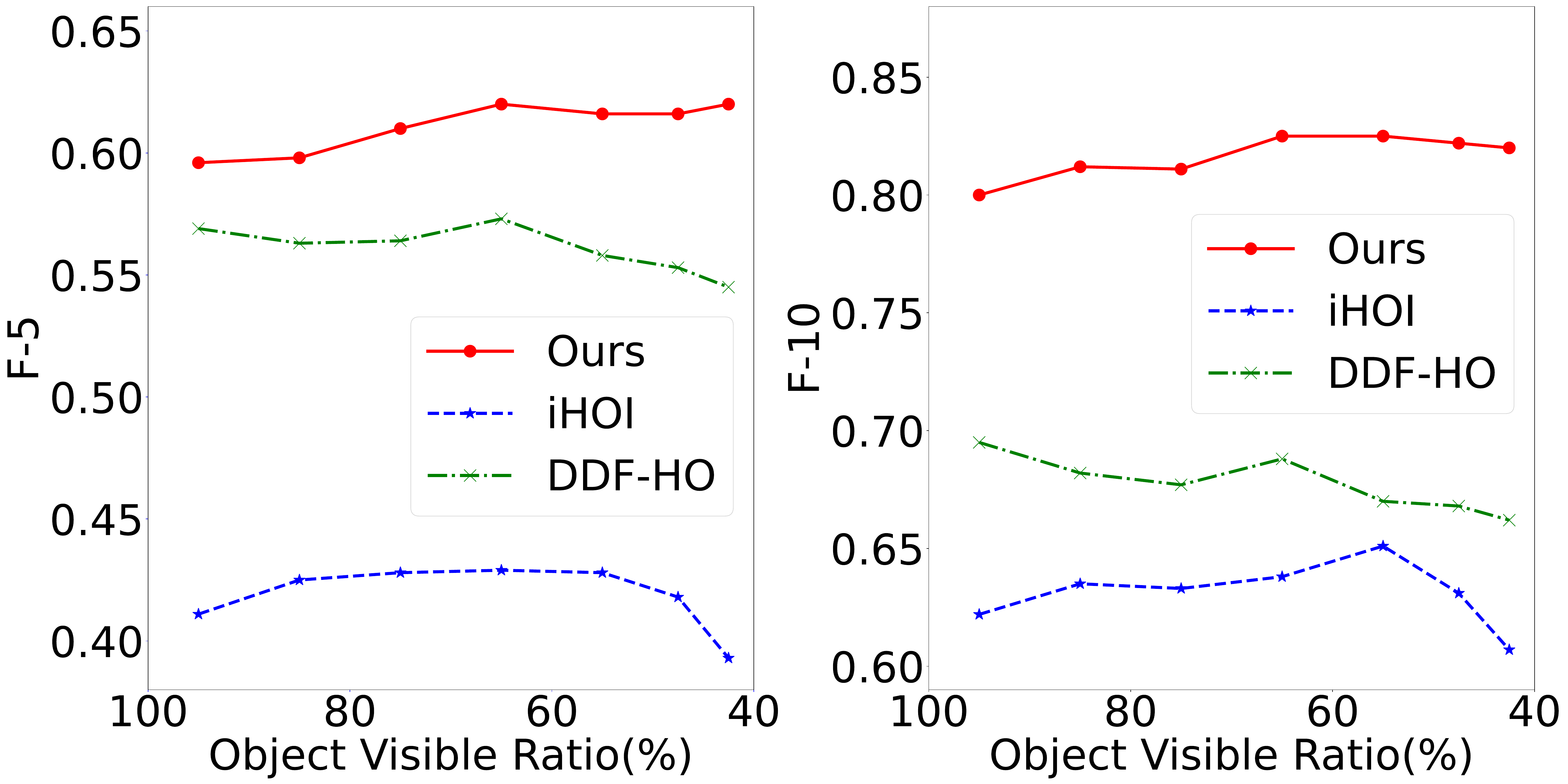}
\caption{\label{fig:ablate_occlusion}}
\end{subfigure}~~
%%%%%%%%%%
\end{table}

%% file: tables/real_finetune_zeroshot.tex
\begin{table}[t]
\centering
\caption{
\label{tab:real_fintune_zeroshot}
{Results on real-world datasets HO3D~\cite{hampali_CVPR20_HO3D} and MOW~\cite{cao2021MOW}. 
We report results for both finetuning (top) and zero-shot transfer (bottom).
}
}
\scalebox{0.88}{
\tablestyle{6pt}{1.0}
\begin{tabular}{c c ccc ccc}
\toprule
\multirow{2}{*}{Method} & \multirow{2}{*}{Finetuning} & \multicolumn{3}{c}{HO3D} & \multicolumn{3}{c}{MOW}\\
 & & F-5 $\uparrow$ & F-10 $\uparrow$ & CD $\downarrow$ & F-5 $\uparrow$ & F-10 $\uparrow$ & CD $\downarrow$ \\
\cmidrule(lr){3-5} \cmidrule(lr){6-8}
HO~\cite{hasson_CVPR19_obman} & \cmark & 0.11 & 0.22 & 4.19 & 0.03 & 0.06 & 49.8\\
GF~\cite{karunratanakul_2020_gf} & \cmark & 0.12 & 0.24 & 4.96 & 0.06 & 0.13 & 40.1 \\
iHOI~\cite{Ye_CVPR22_iHOI} &\cmark & 0.28 & 0.50 & 1.53 & 0.13 & 0.24 & 23.1 \\
DDF-HO~\cite{zhang2024ddf} \footnotemark &\cmark & 0.27 & 0.40 & 0.86 & 0.17 & 0.24 & 1.59 \\
Ours &\cmark & \textbf{0.41} & \textbf{0.63} & \textbf{0.34} & \textbf{0.30} & \textbf{0.47} & \textbf{1.35} \\
\midrule
iHOI~\cite{Ye_CVPR22_iHOI} &\xmark& 0.14 & 0.27 & 4.36 & 0.09 & 0.17 & 8.43 \\
DDF-HO~\cite{zhang2024ddf} &\xmark& 0.24 & 0.36 & \textbf{0.73} & 0.14 & 0.19 & 1.89 \\
Ours &\xmark & \textbf{0.27} & \textbf{0.41} & 1.19 & \textbf{0.24} & \textbf{0.36} & \textbf{1.44}\\
\bottomrule

\end{tabular}}
\end{table}

%% file: tables/dexycb.tex
\begin{table}[t]
\centering
\caption{
\label{tab:sota_dexycb}
{Comparison with the state-of-the-art on DexYCB~\cite{dexycb}. }
}
\tablestyle{7.5pt}{1}
\scalebox{0.88}{
\begin{tabular}{cccccc}
\toprule
Metric & HO~\cite{hasson_CVPR19_obman} & GF~\cite{karunratanakul_2020_gf} & AlignSDF~\cite{chen2022alignsdf} & gSDF~\cite{chen2023gsdf} & Ours\\
\cmidrule(lr){2-6}
F-5 $\uparrow$ & 0.38 & 0.39 & 0.41 & 0.44 & \textbf{0.63}\\
F-10 $\uparrow$ & 0.64 & 0.66 & 0.68 & 0.71 & \textbf{0.82} \\
CD $\downarrow$ & 0.42 & 0.45 & 0.39 & 0.34 & \textbf{0.13} \\
\bottomrule

\end{tabular}
}
\end{table}

%% file: figs/vis_real.tex
\begin{figure*}[t]
    \begin{center}
    \includegraphics[width=0.99\linewidth]{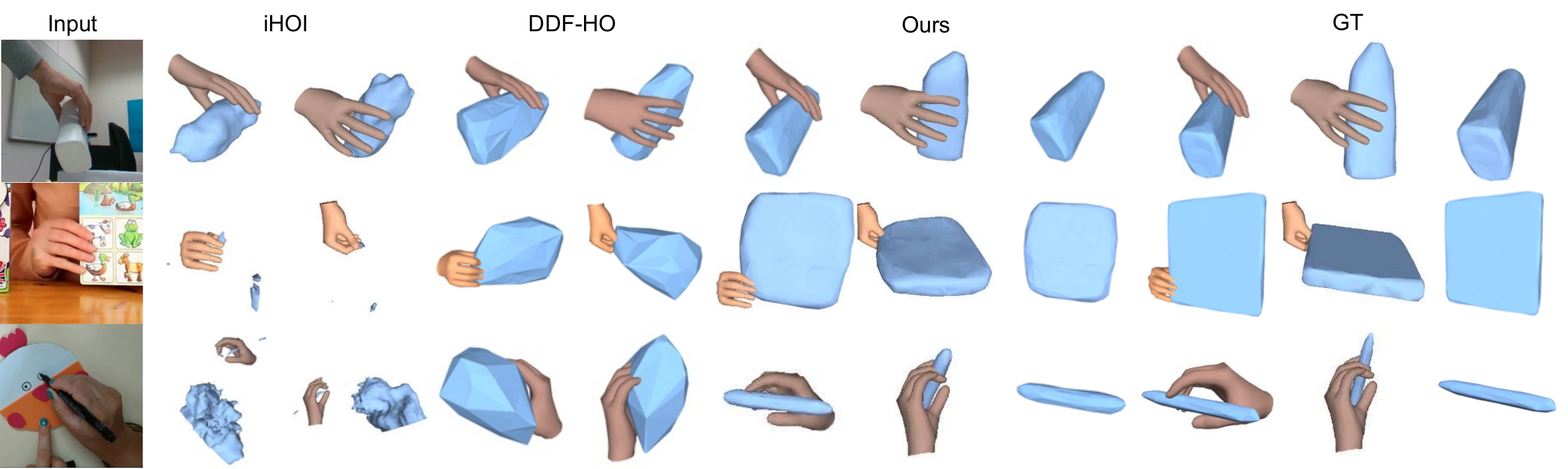}
    \end{center}
    \caption{
        \label{fig:vis_real}
        {Qualitative results on HO3D~\cite{hampali_CVPR20_HO3D} (top) and MOW~\cite{cao2021MOW} (middle and bottom) datasets. For each method and ground truth, we show the reconstruction results in the camera view (column 1) and a novel view (column 2). We also show the unoccluded objects for our method and ground truth. }
    }

\end{figure*}

%% file: tables/ablation.tex
\begin{table*}[t]
\centering
\caption{
\label{tab:ablation}
{Ablation study on ObMan~\cite{hasson_CVPR19_obman} and HO3D~\cite{hampali_CVPR20_HO3D} datasets.}
}
\scalebox{0.88}{
\tablestyle{4pt}{1.0}
\begin{tabular}{cc ccc ccc}
\toprule
\multirow{2}{*}{Row}& \multirow{2}{*}{Method} & \multicolumn{3}{c}{ObMan} & \multicolumn{3}{c}{HO3D} \\
 & & F-5 $\uparrow$ & F-10 $\uparrow$ & CD $\downarrow$ & F-5 $\uparrow$ & F-10 $\uparrow$ & CD $\downarrow$ \\
\cmidrule(lr){1-2}\cmidrule(lr){3-5}\cmidrule(lr){6-8}
A0 & Ours & 0.61 & 0.81 & 0.11 & 0.41 & 0.63 & 0.34 \\
A1 & Ours Oracle & 0.67 & 0.86 & 0.09 & 0.51 & 0.76 & 0.23 \\ 
\cmidrule(lr){1-2}\cmidrule(lr){3-5}\cmidrule(lr){6-8}
B0 & A0 $\rightarrow$ w/o $\mathcal{L}_{mask}$ & 0.57 & 0.76 & 0.23 & 0.36 & 0.56 & 0.61 \\
\cmidrule(lr){1-2}\cmidrule(lr){3-5}\cmidrule(lr){6-8}
C0 & B0 $\rightarrow$ w/o dual-stream denoiser & 0.54 & 0.74 & 0.27 & 0.34 & 0.53 & 0.76 \\
\cmidrule(lr){1-2}\cmidrule(lr){3-5}\cmidrule(lr){6-8}
D0 & C0 $\rightarrow$ w/o $X_t^{HO}$ \& $X_t^A$ & 0.48 & 0.67 & 0.41 & 0.28 & 0.46 & 0.96\\
D1 & C0 $\rightarrow$ w/o $X_t^{HO}$ & 0.51 & 0.69 & 0.37 & 0.33 & 0.50 & 0.81\\
D2 & C0 $\rightarrow$ w/o $X_t^A$ & 0.51 & 0.69 & 0.38 & 0.30 & 0.48 & 0.89 \\
D3 & C0 $\rightarrow$ w/ GCN hand embedding & 0.52 & 0.71 & 0.30 & 0.34 & 0.53 & 0.82 \\
D4 & C0 $\rightarrow$ w/ global hand embedding & 0.52 & 0.71 & 0.30 & 0.31 & 0.49 & 0.86 \\
\cmidrule(lr){1-2}\cmidrule(lr){3-5}\cmidrule(lr){6-8}
E0 & D0 $\rightarrow$ w/o centroid fixing & 0.44 & 0.61 & 0.65 & 0.27 & 0.45 & 1.00\\
E1 & D0 $\rightarrow$ w/o centroid prediction network & 0.32 & 0.45 & 2.48 & 0.23 & 0.36 & 1.31\\
\cmidrule(lr){1-2}\cmidrule(lr){3-5}\cmidrule(lr){6-8}
F0 & E0 $\rightarrow$ Test with GT object centroid & 0.45 & 0.67 & 0.36 & 0.29 & 0.47 & 0.93\\
F1 & E0 $\rightarrow$ Test with GT object pose & 0.50 & 0.70 & 0.34 & 0.31 & 0.49 & 0.84\\
\bottomrule

\end{tabular}
}
\end{table*}

%% file: sec/5_conclusion.tex
\section{Conclusion}
\label{sec:conclusion}
This paper presents \name, a novel centroid-fixed dual-stream conditional diffusion model for single-view hand-held object reconstruction. 
\name does not require any object templates, category priors, or depth information, and excels at modeling uncertainties induced by hand- and self-occlusion. 
In the core, we propose a hand-constrained centroid-fixing paradigm, utilizing estimated hand vertices to prevent the centroid of the partially denoised point cloud from diverging during diffusion and reverse processes. 
Further, a dual-stream denoiser is introduced to semantically and geometrically model hand-object interaction, with a novel unified hand-object semantic embedding enhancing the robustness against occlusion. 
% Our experiments demonstrate that our approach surpasses existing methods in both synthetic and real-world scenarios. 

%% file: sec/6_acknowledgement.tex
\section*{Acknowledgements}
This work was supported in part by the China Scholarship Council and in part by the Shuimu-Zhiyuan Tsinghua Scholar Program. 